\documentclass[letterpaper, 10 pt, conference]{ieeeconf}  

\IEEEoverridecommandlockouts                              

\usepackage{amsmath} 
\usepackage{algorithm}
\usepackage{algpseudocode}
\usepackage{graphicx}
\usepackage{multirow}
\graphicspath{ {./images/} }
\usepackage{letltxmacro}
\LetLtxMacro{\oldalgorithmic}{\algorithmic}
\LetLtxMacro{\endoldalgorithmic}{\endalgorithmic}
\renewenvironment{algorithmic}[1][0]{%
  \hrulefill\par
  \oldalgorithmic}
  {\endoldalgorithmic\par
   \vspace*{-.5\baselineskip}
   \hrulefill\par
  }
  
\title{\LARGE \bf
On CAD Informed Adaptive Robotic Assembly
}

\author{Yotto Koga$^{1}$, Heather Kerrick$^{2}$ and Sachin  Chitta$^{3}$
\thanks{$^{1}$Yotto Koga is with Autodesk Research, San Francisco, CA
        {\tt\small yotto.koga@autodesk.com}}%
\thanks{$^{2}$Heather Kerrick is with Autodesk Fusion 360, San Francisco, CA
        {\tt\small heather.kerrick@autodesk.com}}%
\thanks{$^{3}$Sachin Chitta is with Autodesk Research, San Francisco, CA
        {\tt\small sachin.chitta@autodesk.com}}%
}

\begin{document}

\maketitle
\thispagestyle{empty}
\pagestyle{empty}

\begin{abstract}

We introduce a robotic assembly system that streamlines the design-to-make workflow for going from a CAD model of a product assembly to a fully programmed and adaptive assembly process. Our system captures (in the CAD tool) the intent of the assembly process for a specific robotic workcell and generates a recipe of task-level instructions. By integrating visual sensing with deep-learned perception models, the robots infer the necessary actions to assemble the design from the generated recipe. The perception models are trained directly from simulation, allowing the system to identify various parts based on CAD information. We demonstrate the system with a workcell of two robots to assemble interlocking 3D part designs. We first build and tune the assembly process in simulation, verifying the generated recipe. Finally, the real robotic workcell assembles the design using the same behavior.

\end{abstract}

\section{INTRODUCTION}

Traditionally, automated assembly processes are implemented using fixed automation, requiring considerable effort for each product to ensure robust and reliable execution. However, the needs of ever more demanding and discerning customers are pressuring manufacturers to provide customizable product offerings. As a result, product cycles are shortening, and manufacturers must often retool and re-program their assembly lines to automate production for \emph{high-mix, low volume} products. However, frequent assembly workcell changes are cost-prohibitive because of the required time, material, and programming. Simplifying the programming workflow for automated assembly will lead to a new generation of easy-to-deploy and flexible automation systems. 

This paper describes our efforts toward simplifying the design-to-make workflow using robots through a novel mix of planning, perception, and interactive authoring tools. The interactive tools are designed to allow automation designers to easily specify desired actions at a high level, using their expertise to augment the automated capabilities of our system - {\em e.g.} the designer can specify the assembly sequence order of parts while planning techniques are used to find intermediate waypoints for the robot putting the parts into the assembly.

Through the interactive tools, the designer further simplifies the assembly problem into three separate robotic manipulations tasks, picking up the part, re-grasping the part to obtain the required goal grasp for part insertion, and finally inserting the part into its goal location in the assembly. 

We use depth images from an RGBD camera capturing a collection of parts to feed a fully convolutional DenseNet model to identify part classes and to infer collision-free grasp locations on the parts. This provides automated part picking and adaptation to variations in the part locations. We also feed depth images of grasped parts and the assembly with another DenseNet model (trained differently than the grasping model) to infer 6 DOF pose estimates. Combined with a simple graph-based regrasp sequence planner, this provides automated re-grasping of parts to the desired goal grasp. With pose estimates of the grasped part and the assembly area, the robot can deliver the part through its waypoints into the assembly. We demonstrate our approach by automating a challenging assembly of a set of interlocking part designs with a multi-robot workcell.

Finally, our system makes extensive use of CAD information: (1) In the interaction tool to specify high-level input for the assembly process, (2) In the perception pipeline to enable part manipulation based solely on CAD data of the part for training DenseNet models, and (3) In the planning pipeline to simulate and validate feasible grasps, assembly paths and re-grasping behaviors. We believe a CAD to assembly pipeline of this form will improve the accessibility of assembly automation systems to non-expert users.

In summary, the contributions in this paper are: (1) A novel end-to-end design-to-adaptive-make workflow augmenting designer input with planning and perception, (2) A system that can deal with uncertainties arising from part presentation, from pickup to re-grasping to insertion, (3) A precise pose estimator that allows the robots to assemble the parts of the design using only positional control, and (4) A demonstration of the system on a challenging assembly task with a dual-arm robotic workcell.

\section{RELATED WORK}

Using CAD information for inferring the assembly sequence has been studied in early work \cite{c1,c2} where the concept of assembly by disassembly was used. This approach draws on the fact that part motions are significantly constrained in the disassembly phase, reducing the complexity of finding a solution. Other work inferred feasible assemblies using geometric descriptions and constraints of the parts \cite{c16,c17}. Knepper {\em et al.} built on this approach in \cite{c3} to coordinate a team of robots building furniture.

In \cite{c4}, Michniewicza {\em et al.} outlined the development of a system that combined task and functional primitives into a directed graph called an Augmented Assembly Priority Plan. The plan can consider constraints ({\em e.g.}, workspace, collision) and allocate resources ({\em e.g.}, robots, conveyors, grippers, etc.). Individual operations (called skills – {\em e.g.}, fastening, insertion, etc.) are sequenced appropriately to complete the assembly.

We rely on the designer to provide the assembly sequence order and allocation of resources for assembly tasks.

In \cite{c6}, Lozano-Pérez {\em et al.} described a robot manipulation system called HANDEY. They integrated vision, path planning, grasp planning, regrasp planning, and motion control to pick, regrasp, and place shapes using multiple robot arms. Re-grasping of parts was done by placing them on a table. In \cite{c7}, Alami {\em et al.} presented a unified framework for planning manipulation tasks for picking, re-grasping, and placing objects using a robot arm. In \cite{c8}, Koga and Latombe developed a manipulation planner for multi-arm robotic systems. They demonstrated simulated results of multiple arm robotic systems with up to 24 degrees of freedom, grasping an object, transferring it to some goal location, and initiating re-grasping motions as needed to complete the task. We are similar in spirit to the HANDEY system and decompose the assembly problem into separate tasks, in contrast to the unified and challenging task and manipulation planning approach.

In \cite{c22}, Wan {\em et al.} described a regrasp planning component for finding sequences of robot poses and grasp configurations to reorient a part to some goal pose. They construct the problem as a graph search problem. Our re-grasping approach is very similar, except that we search the graph to find a sequence to reach a goal grasp of the part.

Deep learning models to infer grasp location on parts from RGBD images is a popular method for robotic part picking tasks \cite{c18,c10,c20}. In particular, Redmon and Angelova \cite{c18} described a model to regress RGBD image data into 2D grasp proposals represented by a grasp rectangle $(x, y, \theta, h, w)$. Our grasping model is a simple extension of \cite{c18} to 3D grasp proposals. We replace their model backbone with the DenseNet segmentation model \cite{c12} and use dense prediction from a depth image of part collections (presented as a heightmap) to grasp rectangle point clouds. The inferred grasp rectangle point cloud is the 3D grasp proposal for the associated part. 

For pose estimation of objects, there are several techniques to employ such as registration using ICP \cite{c14}, PointNetLK \cite{c3}, or PVN3D \cite{c21}. Our pose estimation task is for non-occluded single objects, so PointNetLK, PVN3D, and similar pose estimation models are excessive for our needs. ICP has challenges with local minima and slow running speeds. For our pose estimation needs, we modified our training data pipeline for the DenseNet grasp proposal model to produce a pose rectangle point cloud label associated with the view of the part. Given the offset between the pose rectangle and the part frame, we extract a reliable pose estimate of the part after PCA.

In \cite{c23}, Gorjup {\em et al.} described the system they developed for the IROS 2019 Robotic Grasping and Manipulation Competition. The competition consisted of four representative classes of assembly tasks, fastener threading, insertion, wire routing, and belt threading and tensioning. A CAD file described the tasks' layout, ground truth pose, and goals. Their first place system consisted of a CAD interface to extract the ground truth data, a specialized gripper to handle multiple tasks, and compliance control to recover from calibration errors. In \cite{c24}, Drigalski {\em et al.} described the system they built for the World Summit 2018 World Robot Challenge. The set of tasks for this competition was similar to \cite{c23}. Likewise, we have created workflow and tools to simplify the task of setting up automated adaptable robotic assembly of designs. We incorporate many of the same components, although we tackle a more unconstrained problem where parts are arbitrarily presented to the system, which requires solutions for tilting the gripper fingers to pick up the parts successfully, and re-grasping of parts. We also only require positional control to assemble the design.

\section{APPROACH}

For this work, the assembly tasks we consider require picking up parts, re-grasping if necessary, and then inserting parts into the assembly without fasteners like bolts or clips. We consider only parallel-jaw gripper fingers. We assume there are at least two robots in the workcell with a common workspace where they can pass the part between each other for re-grasping. We demonstrate the system using two robot arms to assemble interlocking block designs called Yinan blocks, named after their inventor \cite{c9}.

In this section, we first describe the authoring steps to specify the high-level instructions for the assembly process. We then describe our method of using planning techniques and perception models to fill in the details for the actual robots in the workcell to complete the assembly.

\begin{figure}[thpb]
  \centering
  \includegraphics[scale=0.4]{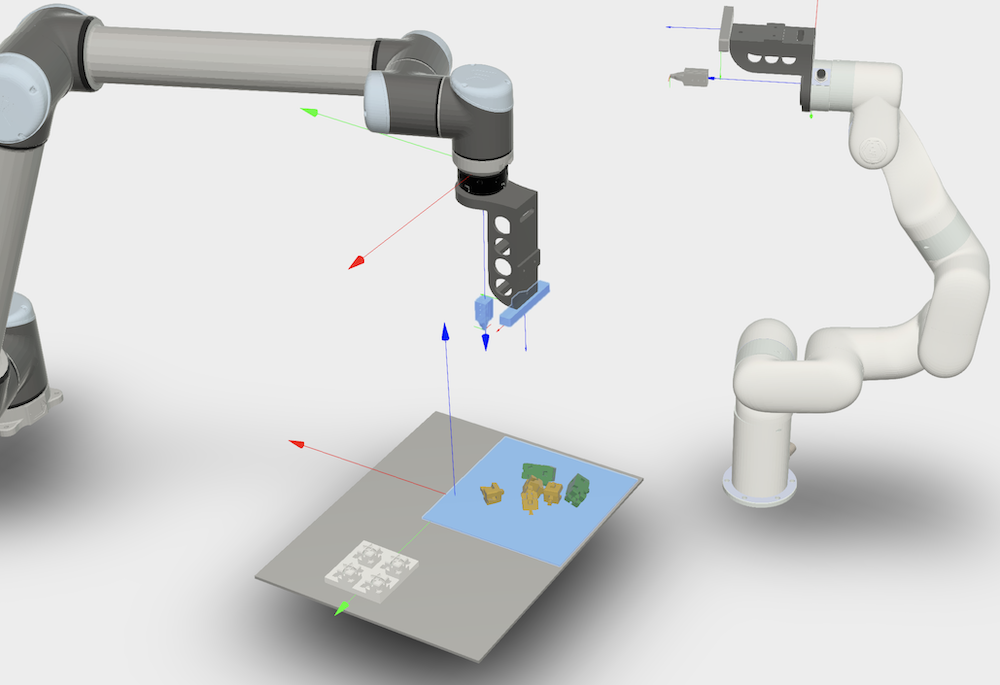}
  \caption{A digital twin of a physical robotic workcell with custom made fingers and camera fixtures. The authored pickup area and camera and fingers used to grasp the parts are highlighted in blue.}
  \label{fig:twin}
\end{figure}

\begin{figure}[thpb]
  \centering
  \includegraphics[scale=0.45]{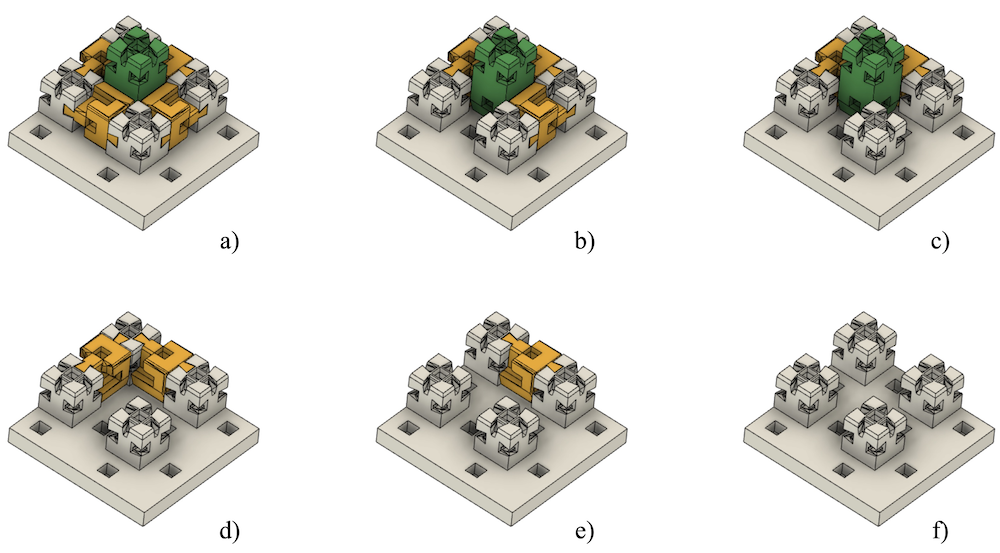}
  \caption{A disassembly sequence for a design made with Yinan blocks. Sequence goes for a) to f).}
  \label{fig:assembly}
\end{figure}

\subsection{Authoring Steps} 
\begin{figure}[thpb]
  \centering
  \includegraphics[scale=0.4]{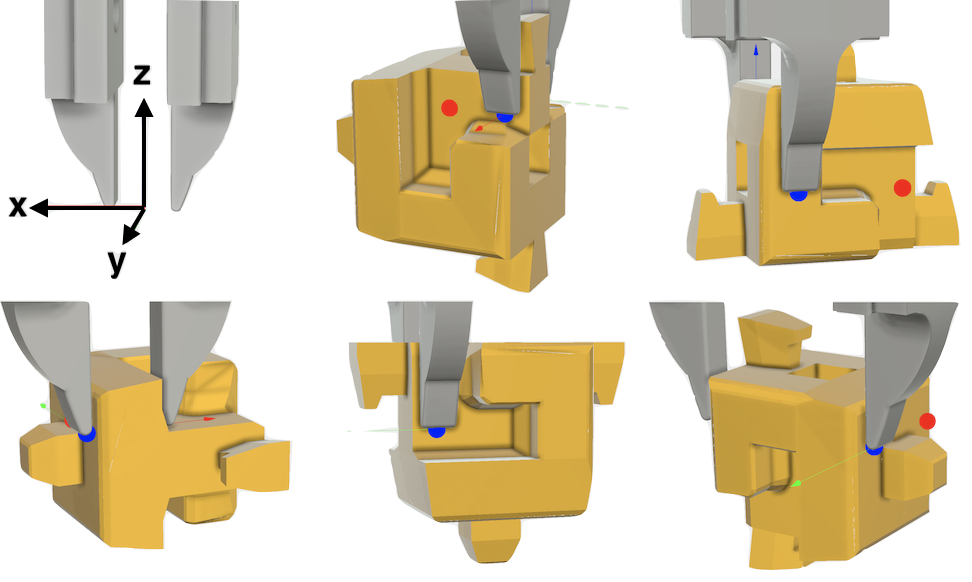}
  \caption{Examples of authored grasps. The upper left cell shows the reference frame for the fingers. The frame origin is at the midpoint between the two fingertips.}
  \label{fig:grasps}
\end{figure}

\begin{figure}[thpb]
  \centering
  \includegraphics[scale=0.4]{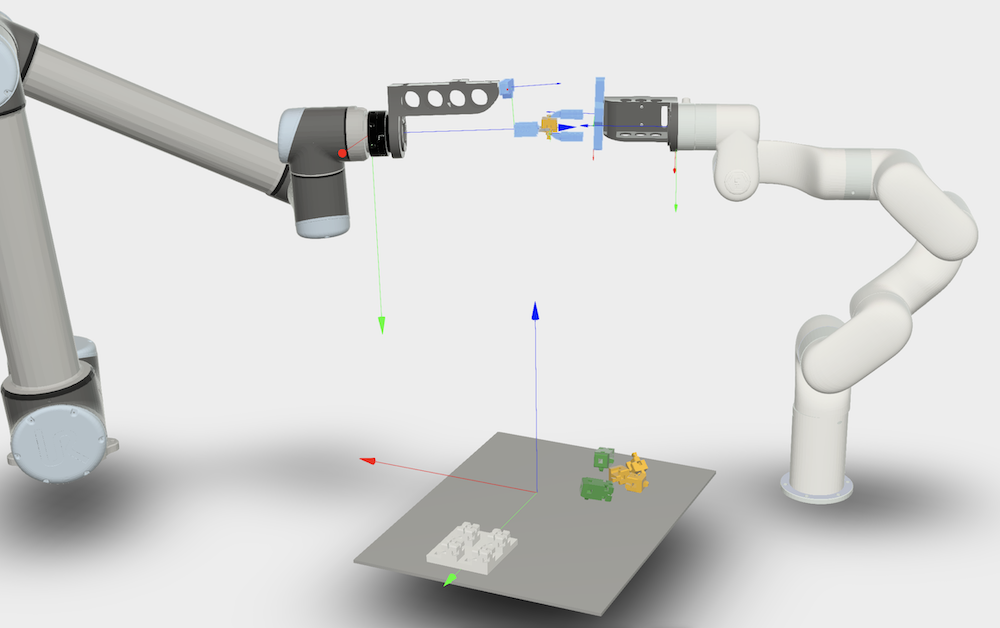}
  \caption{An example of the location and the specified cameras and fingers used to drive regrasp actions of parts (highlighted in blue).}
  \label{fig:regrasp}
\end{figure}

Our approach starts with the designer specifying high-level steps for the assembly process. We assume that the robotic workcell is modeled and available to the designer as a {\em digital twin} (see Fig.~\ref{fig:twin}). The authoring steps of this workflow are the following:
\begin{enumerate}
\item In the CAD tool, the designer specifies the disassembly sequence for the parts and sub-assemblies using a simple point and click visual interface (see Fig.~\ref{fig:assembly}). The authoring tool hides the selected parts to facilitate the specification of the disassembly sequence. Reversing this sequence yields the assembly sequence.

\item The designer authors a set of grasps for each part using the specific fingers in the workcell. A grasp in this set can be a single position and orientation of the fingers with respect to the part or a range between two endpoints. We rely on the designer’s judgment to author stable grasps of the part with the fingers of the gripper. We assume an authored grasp can be flipped by 180 degrees around the fingers' z-axis (see Fig.~\ref{fig:grasps}) and still be valid, so each authored grasp adds two entries into the set. An example of authored grasps in a set is shown in Fig.~\ref{fig:grasps}.
\item The designer can additionally specify locations for the parts (pickup area), the cameras available in the workcell, the assembly location in the workcell, a re-grasping area, and other workcell parameters, including choice of cameras and robots for different parts of the assembly process (see Figs.~\ref{fig:twin} and~\ref{fig:regrasp}).  
\item In the final step, the designer stitches the assembly behaviors together following the sequential task-level goals. We provide task-level APIs to simplify the scripting task. 
\end{enumerate}

The authored steps describe the high-level intent of the assembly process. At each stage, a validation check is run to verify that the assembly sequence, with the robot, their fingers, and cameras in the workcell can reach, insert and view the parts in the specified areas in a collision-free manner. Upon verification of all steps, a recipe of sequential task-level goals for assembling the design is generated. For each part, the instructions include its type, its general location in the workcell, which camera to use to locate the part, the fingers that will grasp it, the camera which will be used to get its pose estimate, the goal grasp to insert it into the assembly, and finally, the intermediate waypoints to insert it into the assembly. Our system requires the designer to reauthor steps due to failures in the validation stage. Incorporating designer feedback in this manner allows our system to draw on the skills and experience of the designer in the areas that they are most skilled at while providing an additional checkpoint before executing the behaviors in the robot workcell. 

\subsection{Robot Steps}

At each stage of the assembly process, parts need to move from the pickup area to the assembly area and be inserted into the assembly. This is a manipulation planning problem \cite{c7,c8} which aims to find the robot motions to grasp the part and transfer it to the goal location in a collision-free manner. Reaching the goal may require the re-grasping of the part using stable placement areas or an extra gripper to hold the part. We simplify the problem by breaking this end-to-end consideration into three separate tasks, 1) grasping and pickup of the part, 2) insertion of the part into the assembly, and if necessary, 3) re-grasping of the part to the required grasp for part insertion.

To pick up the part, we use a perception model to find the required part in the pickup region and infer its collision-free grasp for the user-specified robot. We then move the robot and gripper to this location to grasp and pick up the part. For part insertion, we use a motion planner with grasp constraints for the user-specified insertion robot to find the required grasp of the part for insertion and its waypoints into the assembly. We then move the robot and the grasped part (with the goal grasp) through the waypoints into the assembly. If the goal grasp differs from the pickup grasp, we first move the grasped part into the user-specified regrasping area. A regrasp planner is used to find the sequence of re-grasping steps to achieve the required goal grasp for the insertion robot. We then move the robots through the re-grasping steps. We now describe the key components in more detail.

\subsubsection{Grasp Inference}

We want our system to be able to pick up tilted parts from a pile using a parallel-jaw gripper.
This requires finding grasps that orient in 3D and account for tilt to maximize the chance for successful grasps. 

\begin{figure}[thpb]
  \centering
  \includegraphics[scale=0.49]{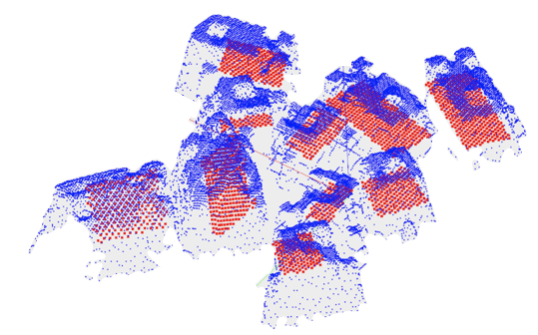}
  \caption{A part pile heightmap (blue points) and the associated 3D grasp proposals (red points). The blue points are the input to the model, and the red points are the output.}
  \label{fig:graspdata}
\end{figure}

Our grasp perception model is a simple extension of the work from Redmon and Angelova \cite{c18}. We replace their model backbone with the DenseNet segmentation model \cite{c12} and use dense prediction from a depth image of piled parts (presented to the model as a heightmap) to get 3D grasp proposals for the parts. A 3D grasp proposal represents the stable and collision-free grasp region of the associated part for the fingers of the robot. The model produces the proposal as a rectangular point cloud (see Fig.~\ref{fig:graspdata}). For each rectangular point cloud, we use PCA to get its 3D center, 3D orientation, width, and height of the grasp proposal. The center and normal vector for the rectangular region are the proposal frame origin and z-axis direction, respectively. The points in the proposal are located at the pixel coordinates of the dense prediction. The height value at the output pixel is the height of that point in the rectangular point cloud. Also, the part class id of the associated part is concatenated by the model to each pixel height value.

\begin{figure}[thpb]
  \centering
  \includegraphics[scale=0.49]{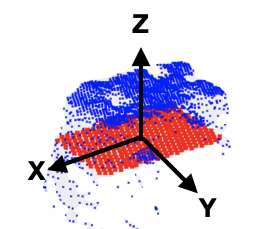}
  \caption{The grasp proposal frame associated with a heightmap of a part.}
  \label{fig:graspproposal}
\end{figure}

\begin{figure}[thpb]
  \centering
  \includegraphics[scale=0.49]{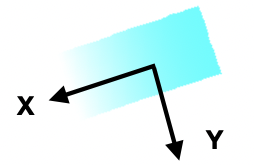}
  \caption{The grasp x-gradient.}
  \label{fig:graspgrad}
\end{figure}

The frame of the grasp proposal (see Fig.~\ref{fig:graspproposal}) defines the location of the finger frame for grasping (see Fig.~\ref{fig:grasps}). From just the rectangular point cloud, the direction for the fingers to close on the part is ambiguous. Is it along the width or height of the proposal? A value gradient over the proposal gives this direction and the model concatenates gradient samples to the pixels' height and class id values (see Fig.~\ref{fig:graspgrad}). The x-axis direction for the grasp proposal (the finger closing direction) is along the gradient going from biggest to smallest value. 

The training data is generated synthetically by randomly placing and piling the CAD model of the parts in the pickup area in different orientations. We use a 3D OpenGL-based renderer to create the depth data of the parts from the view of the perspective camera for a given sample. To facilitate the transfer of the learned model to work with actual camera depth data, we add noise to the rendered depth map by downsampling the resolution and adding Perlin noise to its depth values. We convert the depth image into an orthographic heightmap representing the parts' structured 2.5D point cloud.

The associated grasp proposal label for the sample is computed in the following manner. For each part in the pile, the CAD model of the grasping fingers is positioned at each grasp in the set authored for that part. A grasp is considered valid if the CAD model of the fingers grasping the part does not collide with any other parts or the environment. When a grasp defines a range between two endpoints of the fingers along the part, we discretize the range into a few steps and check if each step is valid. Consecutive, valid tests for a grasp are grouped to create the grasp proposal. The grasp proposal frame is the fingers' tip frame at the midpoint of the span. Since the finger frame can be rotated by 180 degrees around its z-axis and be the same grasp, we avoid ambiguity in the label by constraining the proposal frame to have its x-axis direction always pointing towards the right half of the camera view. The dimension of the proposal in its x-axis is the spacing between the fingers for the grasp, and the dimension in its y-axis is the finger width plus the span of consecutive, valid tests. For all valid proposals, we check for any overlap between proposals from the camera view. If there is an overlap, we discard the bottom proposal. We then render the remaining proposals into a heightmap that matches the associated part heightmap sample (rendered from the same camera viewpoint). We save the label in an image format. The height of the grasp proposal pixels are encoded in the alpha channel of the rendered result. The grasp proposal x-axis direction gradient is encoded in the red channel. And finally, the part class associated with the proposal pixels are encoded in the green channel.

For each set of parts in an assembly process, we synthesize roughly 600,000 samples of these parts and their associated labels.  Our loss function used in training the DenseNet model is the sum of the cross-entropy loss of the height, class, and gradient value in the dense prediction. We scale the loss by the number of points with non-zero height values in the label to account for the sparseness of the proposals. Our data generation pipeline creates training data at a fixed camera distance from the part pile and generates the associated grasp region labels with a specified finger CAD model. The model take about 12 hours to train on an Nvidia V100 GPU.

We make grasp inference by capturing a heightmap from the specified camera for part pickup and then feeding it to the model. We use 2D clustering techniques to isolate each grasp proposal in the result. We then use PCA on the points to get the center and tilted orientation of the proposal in the camera frame. The part class associated with the proposal is the max class value. The x-axis direction for the frame comes from the gradient direction embedded in the proposal and is updated accordingly.

To grasp a particular part by class, we look at the grasp proposals with the desired part class and choose the one with its center closest to the camera. The grasp pose is mapped to the workcell frame, the fingers open to the value given by the grasp proposal (plus some extra padding), and the robot moves the fingers into position to grasp and pick up the part.

\subsubsection{Motion Planning for Assembly}
\begin{figure}[thpb]
  \centering
  \includegraphics[scale=0.4]{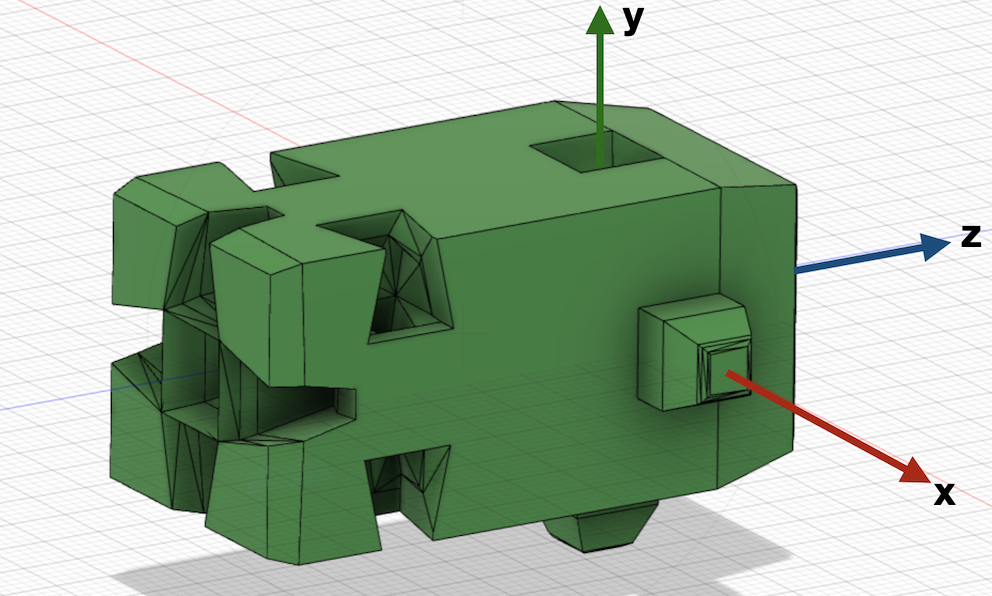}
  \caption{A part designed in a CAD tool where the primary motion directions for insertion into the assembly are also the axes of its local frame (the red, blue, and green arrows).}
  \label{fig:partaxes}
\end{figure}

After the designer authors a disassembly sequence, we use a motion planner to verify that the disassembly sequence is correct. We approximate the test by running a best-first-search \cite{c13} with three translation degrees of freedom along the axes of the part local frame (see Fig.~\ref{fig:partaxes}). This requires that the part axes are aligned with the directions in which the part comes out of the assembly. This seems to be a reasonable requirement regarding how CAD models are created in practice. This allows us to orient the part in its final pose and discretize the translation search space around the tight thresholds of the parts in the assembly and maximize the chance of finding collision-free paths in the discretized configuration space. If a collision-free path to extract the part (to some free space outside the assembly) cannot be found, then an error is communicated to the designer that the sequence is incorrect. We require the designer to re-author the disassembly sequence to correct the failure. Parts that need twisting or tilting into the assembly will require a more exhaustive search routine. In complicated cases, we would have the designer author the waypoints. Despite this limitation, we were still able to explore interesting assembly tasks.

We also add a grasping constraint to the motion planning problem. The grasp constraint requires that the robot and gripper fingers maintain the grasp of the part during the search. The part, robot, and fingers must also be collision-free throughout the path. The disassembly sequence for the design is verified if we can find a collision-free extraction motion and meet the grasp constraint criteria for all parts, running the planner multiple times with the entire design in different random starting poses in the assembly area. If multiple grasps exist, we choose the grasp with the fingers' x-axis aligned with any insertion motion. This is the stable grasp to keep the part from twisting between the fingers during contact with the insertion surfaces.

The grasp associated with the resulting collision-free path is the goal grasp. The motion planning search is biased to find straight-line motions with the minimum number of direction changes. The ends of the straight-line motion become waypoints for the disassembly of a part. The reverse order of waypoints becomes the robot's insertion instructions to move the part into the assembly. For example, the insertion waypoints for the block in Fig.~\ref{fig:partaxes} in the assembly shown in Fig.~\ref{fig:assembly}, is a downward motion with the nub of the part moving through the vertical slot and then sliding horizontally through the horizontal slot embedded inside the assembly to its goal.

Note that we only use the motion planner to find the part’s goal grasp and generate the waypoints to deliver the part to the assembly from just above the assembly area. We use simpler straight-line motions of the fingers or cameras for gross robot motion in the workcell. For uncluttered workcells, this seems sufficient to generate collision-free trajectories. As such, gross motion planning was not a focus of this work, but, in more cluttered workcells, we aim to integrate standard motion planning techniques in the future ({\em e.g.}, using open-source packages like MoveIt! \cite{MoveIt}). 

\subsubsection{Re-grasp Planning}

\begin{figure}[thpb]
  \centering
  \includegraphics[scale=0.49]{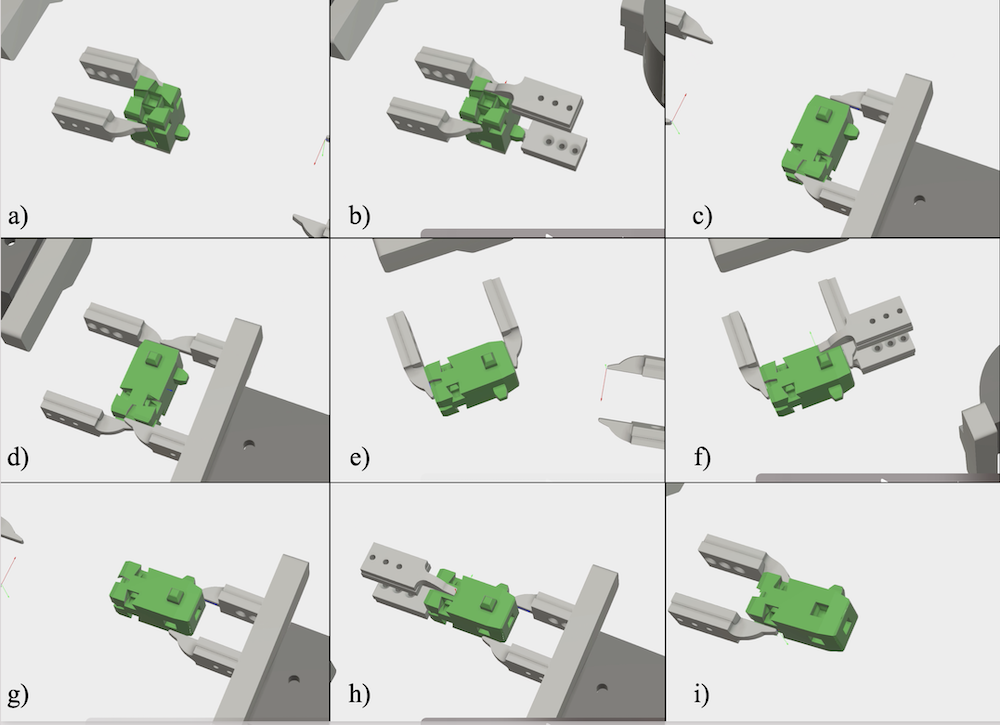}
  \caption{An example re-grasping sequence of a part with the two robot fingers to achieve the goal grasp. Sequence goes from a) to i).}
  \label{fig:regraspseq}
\end{figure}

The parts in the pickup area can have arbitrary positions and orientations. The grasp chosen for pickup may differ from the required grasp for part insertion necessitating a re-grasping step. To re-grasp the part, we use two robots to move through a sequence of alternating grasps until the desired grasp is achieved (see Fig.~\ref{fig:regraspseq}). 

We cast the problem into a graph search to find the re-grasping steps. We construct the graph by first specifying where re-grasping occurs in the workcell. The part will be centered at this location but can have different orientations. We require a finite set of varying part poses to seed the graph. For re-grasping, one robot will always hold the part. We compute the associated grasp pose for the given part pose for each grasp definition in the grasp set. We test if this grasp pose for the fingers is achievable. We then test if the robot and fingers are collision-free with the environment for achievable grasps. If these conditions are met, this $(grasp, part pose, finger)$ tuple becomes a node in our graph. To manage grasp definitions that allow the fingers to be in a range along an edge of a part, we break up the range into three samples, the grasp at two endpoints along the part and the midpoint.

Two nodes of the graph are connected when they share the same part pose, the fingers in the nodes are different (i.e. the robots in the nodes are different), and the robots, fingers, and any attached peripherals to the robots grasping the part simultaneously do not collide with each other (see Fig.~\ref{fig:regrasp}). This edge type is a re-grasp step. Nodes with the same grasp and fingers are also connected if the fingers (and associated robot) can move the part from the pose in the first node to the second node in a collision-free and reachable manner. This edge type is a repose step. 

Assuming we can determine the initial grasp definition of the part with the fingers, we find the nodes in the graph with the same grasp, finger pair. For each node in this set, we can traverse edges of the graph using best-first search and find the minimal number of re-grasp steps to achieve the desired goal grasp of the part if a path exists. The path with the minimum number of re-grasp steps is chosen if multiple paths exist. 

To get the initial grasp definition of the part after pickup, we first get its pose estimate. This gives us an estimate of the offset between the grasping fingers' frame and the part. We then visit the nodes in the graph with the free fingers (the fingers not grasping the part). For the associated part pose, we determine if the grasping fingers can repose the part to that pose and that the grasp for the free fingers will be clear of the current fingers. For nodes that pass this test, we search the graph to find the node that yields the shortest re-grasp sequence. We then re-grasp the part according to that node. This gives us the initial grasp definition of the part.

For re-grasping the part, we first move the part into the shared part pose associated with the edge using the fingers currently grasping the part. Next, we move the free fingers in a linear motion to a hovering position just above its grasp pose with the fingers opened slightly greater than its grasp definition. Then, we move the fingers in a linear motion to the grasp pose and close the fingers around the part. The other fingers release the part, and we move it in a linear motion just above its last grasp. Finally, we move the fingers to an area away from the part to give clearance for the reorienting part. We repeat this process until the goal grasp is reached.

\subsubsection{Part Pose Estimation Model}

We need a precise in-hand pose estimate of the grasped part for the robot to accurately deliver the fingers to their re-grasp locations and the part to its insertion location. Note that the pose estimate only needs to be precise since we can rely on the precision of the robots and calibration to deliver the fingers and the part accurately to their goal location. 

\begin{figure}[thpb]
  \centering
  \includegraphics[scale=0.35]{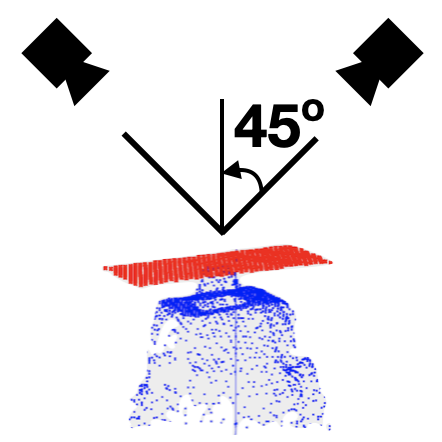}
  \caption{The pose proposal (red points) associated with the heightmap of the part (blue points). This pose proposal is for the viewing directions from the top side of the part. }
  \label{fig:posedata}
\end{figure}

We re-use the DenseNet grasp proposal model and change the training dataset to have the model produce the part pose estimate. The basic idea is to feed the model a point cloud of the part  (presented as a heightmap) and have it produce a rectangular point cloud attached to the part for the given view direction (see Fig.~\ref{fig:posedata}). We call this rectangular point cloud a pose proposal. We extract the part pose estimate from the pose proposal in the following manner.

We partition any view directions of the part into a top, bottom, left, right, front and back direction. These correspond to the six primary viewing directions of the part along the frame axes of the part. A view direction of the part that is within a 45-degree frustum of a primary direction is said to have that primary view direction. For example, in Fig.~\ref{fig:posedata}, all views of the part within 45 degrees of the top side direction are considered a top view direction.

For each primary view direction, there is one associated pose proposal. The model produces the pose proposal for the view direction of the input part heightmap. Each pose proposal has a fixed offset from the part frame, so using PCA to get the center and 3D orientation of the proposal (in the exact manner as grasp proposals) and applying the offset, we can infer the pose estimate of the part. 

The synthetic training data is generated by sampling a heightmap of each part randomly positioned in front of the camera. The distance between the parts and the camera is controlled in a fixed interval that ensures all parts are fully viewable by the camera and fills roughly a third of the view. The height map samples are generated using the same renderer as the grasping dataset.

The associated pose proposal label for the sample is computed as follows. We first create the pose proposal for each primary view direction for each part. This is done by taking the bounding box of the part and making each face the pose proposal for the associated view direction. For example, the bounding box face on the top side is the pose proposal for the top view direction. The frame for each pose proposal is centered on the face, with the x and y axes aligned with the width and height directions and the z-axis along the face normal. The width and height of the face are the width and height of the associated pose proposal. We record the offset between the part frame and each of the six pose proposal frames. We then determine the primary view direction for the given heightmap sample of the part. For that view direction, we take the associated pose proposal, position it with respect to the part frame and then render the rectangular pose proposal into a heightmap that matches the associated part heightmap sample. The height of the pose proposal pixels are encoded in the alpha channel of the rendered result. The pose proposal x-axis direction gradient is encoded in the red channel. And finally, the part class and primary view id associated with the proposal pixels are encoded in the green channel. The combined id is:  $id = view id + 6*part id$.

Some view directions of the parts have rotational symmetry order greater than 1. For example, each view direction of a cube shape will have rotational symmetry of 4. We handle this ambiguity by constraining the direction of the x-axis of the pose proposal during data generation. For example, a view direction with rotational symmetry of order 2, we constrain the x-axis direction of the proposal to point towards the right half of the camera view. 

For each set of parts in an assembly process, we synthesize roughly 400,000 samples and their associated labels.  We use the same loss function from the grasping model. The model take about 12 hours to train on an Nvidia V100 GPU.

We estimate the pose by capturing a heightmap of the part at roughly the same camera distance as in the training set. We use PCA on the points of the generated pose proposal to get the center and tilted orientation of the proposal in the camera frame (the x-axis direction is determined from the gradient values). The part class and view direction id associated with the proposal is the max class value. The offset of the pose proposal to the part frame is retrieved using the part and view direction id. The part pose is then transformed into the camera frame.

\section{EXPERIMENTS}

We implemented the end-to-end workflow using an internal robotics research platform. The authoring and simulation tools are integrated into Autodesk Fusion 360.

\begin{figure}[thpb]
  \centering
  \includegraphics[scale=0.2]{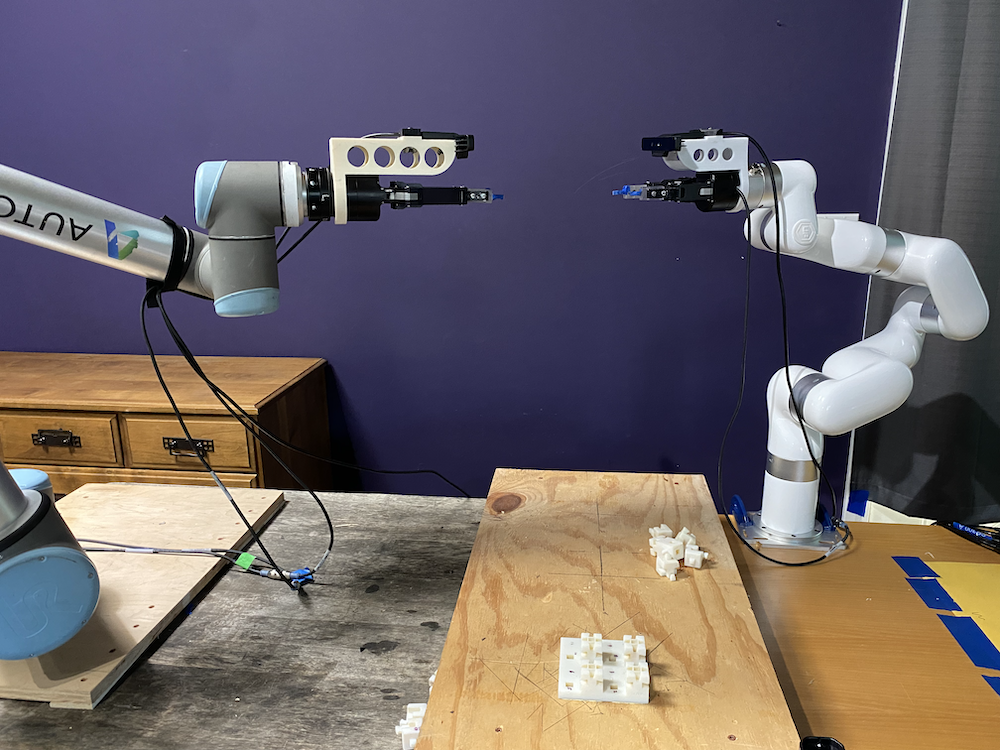}
  \caption{The assembly workcell with two robot arms.}
  \label{fig:realworkcell}
\end{figure}

For the assembly demonstration, we used two shapes which we call Yinan blocks (see Figs. 4 and 8), and a fixed base to support the assembly (see Fig.~\ref{fig:assembly}f). The design we assembled is shown in Fig.~\ref{fig:assembly}a. We calibrated the workcell shown in Fig.~\ref{fig:realworkcell} to create the digital twin shown in Fig.~\ref{fig:twin}. We use a UR10 and a XArm-6 robot, with the Robotiq Rq140 and Rq85 grippers attached to each robot, respectively. The fingers are custom printed for grasping the blocks (see Fig.~\ref{fig:grasps}). We used the Intel RealSense SR-300 cameras for depth sensing. A camera is attached to the end of each robot with a fixture. The Yinan blocks are small (about one in.) and interface with each other along particular pairs of faces. The parts are initially located in random piles in an input area, {\em i.e.}, the parts are not nicely presented to the robot. 

We authored the disassembly sequence for the design by selecting the parts in the order shown in Fig.~\ref{fig:assembly}. The behavior script corresponding to the generated recipe for this experiment is represented in Fig.~\ref{fig:pseudocode}.
\begin{figure}[thpb]
\begin{algorithmic}
\footnotesize
\State $sequence$ = $recipe$.get\_assembly\_sequence()
\State $camera$ = $recipe$.assembly\_camera
\State $^{W}T^{B}$ = $camera$.get\_pose($recipe$.assembly\_area)
\While {$s$ in $sequence$}
    	\State $s$.fingers.pickup\_part($recipe$.pickup\_area, $s$.camera, $s$.part\_class)
    	\State $^{P}T^{F}$ = $s$.fingers.regrasp($s$.goal\_grasp,\\\hspace{3.3cm} $s$.camera,\\\hspace{3.3cm} $s$.other\_fingers, \\\hspace{3.3cm} $s$.other\_camera)
    	\While {$^{W}T^{F}$ in  $s$.insertion\_waypoints($^{W}T^{B}, ^{P}T^{F}$)}
    		\State $s$.goal\_fingers.move\_to($^{W}T^{F}$)
    	\EndWhile

    	\State $s$.goal\_fingers.release()
\EndWhile
\\
\State $^{W}T^{B}$ - fixed base transform in the workcell frame
\State $^{P}T^{F}$ - finger transform in the part frame
\State $^{W}T^{F}$ - finger transform in the workcell frame
\end{algorithmic}
 \caption{Pseudo-code of the assembly behavior for the experiment.}\label{fig:pseudocode}
\end{figure}

\begin{figure}[thpb]
  \centering
  \includegraphics[scale=0.44]{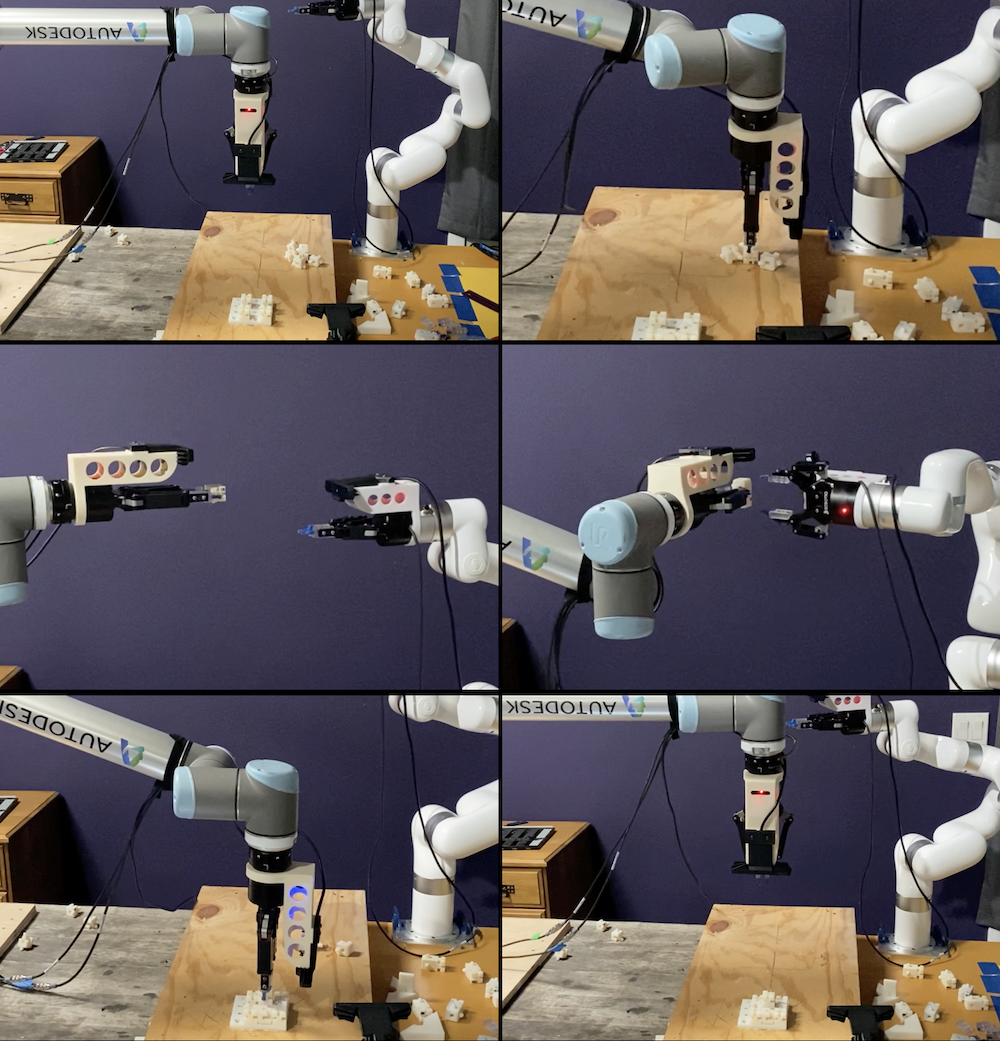}
  \caption{Snapshots from the assembly process. The upper left cell is the first step where the fixed base pose is acquired. The lower right cell is the completed assembly.}
  \label{fig:actual}
\end{figure}

We first ran the behavior in simulation using the digital twin of our workcell to verify that the assembly process was correct. We randomly piled the blocks in the pickup area and then successfully ran the behavior on the actual workcell. To accurately re-grasp and insert the grasped parts into the fixed base (tight tolerances), we had to add a correction map to compensate for the cascading calibration errors between the various fingers, cameras, and robot base offsets. It took 9 minutes for the robots to complete this assembly task (Fig.~\ref{fig:actual}). Inference time for grasp and pose estimation is roughly 50 ms on an Nvidia V100. To capture a clean depth image from the cameras, though, we wait for the robot to stop moving and settle, which adds an extra second to the total inference time. See https://youtu.be/hoAJPZuMmJ0 to view the assembly behavior in action.

The success rate data for the pickup, re-grasp, and insertion tasks are in Table \ref{tab:rate}. We ran six consecutive end-to-end assembly of the design. On failure, we tried the task again by  resuming the assembly sequence from the pickup task. We count each individual re-grasp step. Note that this data was collected with the Xarm replaced with another UR10.

\begin{table}[h]
\caption{Task Success Rate}
\label{tab:rate}
\begin{center}
\begin{tabular}{|c||c|c|c|}
\hline
\bf Task & \bf Success & \bf Failure & \bf Success Rate\\
\hline
Pick & 47 & 2 & 96\%\\
\hline
Regrasp & 130 & 17 & 88\%\\
\hline
Insertion & 30 & 0 & 100\%\\
\hline
\end{tabular}
\end{center}
\end{table}

\section{Discussion}
There are avenues for improvement and further work. The assembly process is slow, and we're working towards improving cycle times for the task. The system is sensitive to the calibration of the robots, their cameras, and each other - {\em e.g.}, failures during re-grasping are when the part is rotated from its pose estimate, and the re-grasping finger has to rotate accordingly. This rotation accentuates calibration errors causing the re-grasping fingers to miss their mark. We plan to use visual servoing to remove the need for the correction maps and the dependency on accurate calibration. We are working towards further automating parts of the workflow that may need input from the designer. This includes determining grasp poses for different parts and motion planning for gross robot motion. We'd also like to improve our grasping and pose estimation models' performance, accuracy, and generality. Currently, our perception models do not generalize to unseen objects. Characterizing the range of shapes our perception models can handle is another area of future work. Our system relies on clean depth images, and we are working towards identifying the right visual sensors for different materials and parts' surface finishes. For insertion tasks that are highly occluded or require tighter tolerances, we aim to explore F/T sensing and insertion policies that could be learned using reinforcement learning techniques.

Assembly sequence planning is another area of interest for further automation. CAD information for most assemblies is often incomplete and inaccurate. Individual parts may not be annotated or labeled correctly, and constraints between parts can often be represented in multiple ways. Furthermore, there is often a mismatch between CAD and real models. We aim to explore these areas using tools from simulation, control, and reinforcement learning. We also aim to explore more complex assemblies in the industrial domain to highlight the advantages of our approach over traditional fixed automation.

\addtolength{\textheight}{-12cm}   





\end{document}